# Few-shot crack image classification using clip based on bayesian optimization

Yingchao Zhang [a] and Cheng Liu [ab*]

[a] Department of Systems Engineering, City University of Hong Kong, Hong Kong, China
E-mails: yingchao.zhang@my.cityu.edu.hk, cliu647@cityu.edu.hk

[b] Centre for Intelligent Multidimensional Data Analysis, City University of Hong Kong, Hong Kong, China

**Abstract**
This study proposes a novel few-shot crack image classification model based on CLIP and Bayesian optimization. By combining multimodal information and Bayesian approach, the model achieves efficient classification of crack images in a small number of training samples. The CLIP model employs its robust feature extraction capabilities to facilitate precise classification with a limited number of samples. In contrast, Bayesian optimisation enhances the robustness and generalization of the model, while reducing the reliance on extensive labelled data. The results demonstrate that the model exhibits robust performance across a diverse range of dataset scales, particularly in the context of small sample sets. The study validates the potential of the method in civil engineering crack classification.

## 1 INTRODUCTION

With the rapid pace of urbanization and the aging of existing infrastructure, the detection and maintenance of structural cracks have become critical to ensuring engineering safety. Cracks not only compromise the aesthetic integrity of buildings and transport infrastructure but also pose risks of structural failure. Such failures can significantly endanger the safety of both individuals and property. Consequently, the timely and precise classification of crack images is crucial for preventing severe accidents and extending the service life of structures.

At present, the most common methods for detecting cracks are manual inspection and digital image processing techniques. However, manual inspection is inefficient and difficult to adapt to large-scale infrastructure inspections. Furthermore, while digital image processing techniques achieve high-precision classification of crack images, the robustness of these techniques is poor and difficult to transfer to cracks on different structures.

The appearance of computer vision technology brings a good solution to the problems of crack classification, detection, and segmentation [1,2,3]. [1] realized high accuracy classification of pavement damage based on inertial measurement unit on vehicles. An enhanced accuracy for detection of crack images captured by UAVs was proposed by [2] through the construction of a multi-level attention mechanism. In contrast, the objective of [3] is to develop a lightweight detection model that can be utilized in edge devices. The model is further optimized by reducing the number of parameters in the baseline model through the implementation of a unique structure, thereby enhancing the inference speed of the model. In addition to crack classification and detection, [4] proposed an algorithm for crack pixel-level classification. This model can determine whether a pixel in images belongs to a crack's pixel or not, thereby further improving the application of computer vision algorithms in civil engineering. To deal with the interference of noise such as tree shadows, [5] used the discrete cosine transform to reduce the interference of noise in images.

Although the above models can achieve varying degrees of crack detection, their F1 values are generally around 0.8. This accuracy is difficult to scale up to large scale infrastructure inspections. Therefore, some advanced models need to be used to advance the application of computer vision algorithms in crack detection.

The main challenges currently faced in the field of crack detection or classification are the following:

1. Low robustness: cracks in different parts of a building have large differences in morphology or colour, which can lead to unstable performance of existing detection and classification algorithms in different scenarios. When it is necessary to classify or identify cracks in a new scene, it is often necessary to retrain the detection model.
2. High dataset demand: current computer vision algorithms require a large amount of labelled data for training. But the acquisition of crack images is very time-consuming. Lack of sufficiently diverse datasets limits the generalisation ability of the model, leading to poor performance in practical applications.

The proposed large vision language model can solve these problems well. The Contrastive Language-Image Pre-Training model (CLIP) [6] is a multimodal model that can combine image and textual information to deepen the model's understanding of the semantic and geometric features of the cracks. The multimodal information helps to improve the accuracy of the model in classifying crack images in different scenes or different structures. CLIP is pre-trained on large datasets with strong generalisation ability, which can be used for zero-shot learning or less-shot learning in the field of crack classification. Since the CLIP model already has some semantic understanding of targets such as cracks, this study will be based on the CLIP model for few-shot learning of crack image classification.

In addition, Bayesian methods can capture model uncertainty and provide greater robustness and generalisation by probabilistically modelling the weights. Bayesian inference can also optimise model parameters with less data, improving model performance.

In this study, we propose a crack image classification model for few-shot learning based on the Bayesian approach with CLIP. The effectiveness of the model is verified through experiments on a small number of crack image datasets.

Section 2 of this paper focuses on the dataset, Section 3 describes the proposed classification model based on the Bayesian approach with CLIP, the results and discussion are arranged in Section 4, and the conclusion is in Section 5.

## 2 CRACK CLASSIFICATION DATASET

We collected images for crack classification from the Roboflow platform, a total of 20,000 images, as shown in Figure 1. We divided them into training set and test set in the ratio of 1:1. To perform less sample learning, we take the images in the training set as the new training set in different ratios, as shown in Table 1. In the training process, the test set is always 10,000 images.

## 3 CLASSIFICATION MODEL

This section introduces Crack-CLIP, a crack image classification model based on CLIP with a Bayesian network. The model comprises two principal components: a CLIP-based feature extraction network and a Bayesian classification network.

### 3.1 CLIP Model

In this study, CLIP was used as the backbone model to achieve effective alignment between images and text. The architecture of the model includes three core components: the text encoder, the image encoder, and its interaction mechanism, as shown in Figure 2. Each component is described in detail below.

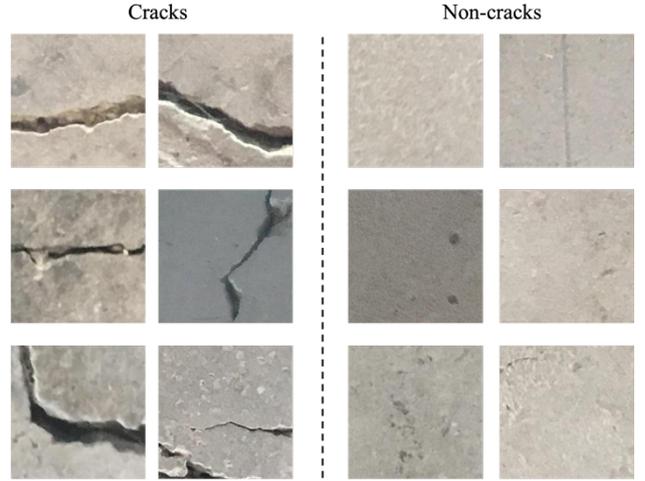

Figure 1: Samples of cracks classification dataset

As shown in Figure 1, the text labels of the two types of samples are "A picture with cracks" and "A picture without cracks".

Table 1: Few-shot training set partitioning

| Number | Ratio (%) | Images Number |
| --- | --- | --- |
| T0 | 0 | 0 |
| T1 | 1 | 100 |
| T2 | 5 | 500 |
| T3 | 10 | 1,000 |
| T4 | 50 | 5,000 |
| T5 | 100 | 10,000 |

#### 3.1.1 Text encoder

The text encoder is mainly used to transform the input text information into a high-dimensional semantic representation. In this study, the text label "A picture without cracks" is transformed into a high-dimensional semantic vector through the following steps by the text encoder.

1. The input text is subjected to the process of tokenization, which generates a sequence of corresponding words, thus producing a format that can be processed by the text encoder.
2. A high-dimensional vector is generated for each vocabulary word through the embedding layer, thereby forming a preliminary word vector representation.
3. These word vectors are fed into the Transformer architecture for further processing to generate context-aware feature representations. These representations effectively capture the syntactic and structural information of the text.
4. The final output of the encoder is a synthesized text feature vector $T$, which is used for contrastive learning with the image features, the expression of which is shown in Equation (1).

$$\boldsymbol{T} = [T_1, T_2, \ldots, T_n] \quad (1)$$

where *n* denotes the length of the text feature vector ***T***, which was set to 512 in this study.

### 3.1.2 Image encoder

The function of the image encoder is to transform the input image into the corresponding image features. In Crack-CLIP model, ViT-B/32 [7] image encoder is used for extracting image features. In order to obtain the final image feature vector ***I***, the input image must proceed through the following steps. The image encoder processing is illustrated in Figure 3.

1. The resolution of the image being input was fixed at 224×224 and subsequently normalised. To ensure consistency and quality of the input data, the pixel values were scaled from [0,255] to [0,1].
2. The input image is sliced into 32×32 patches. An image with a resolution of 224×224 will get 7×7 = 49 patches.
3. Each patch is subjected to a linear transformation, resulting in a 768-dimensional vector. A total of 49 patches are mapped to a 49×768-dimensional tensor.
4. The class token and the location information are also added to the tensor obtained previously.
5. A total of 12 Transformer structures are employed for the processing of the feature maps. The dimensions of the input and output feature maps are maintained throughout the process.
6. The output feature map is linearly transformed to yield a 1×512 image feature. This shape is consistent with the text features as previously described.

### 3.1.3 Interaction mechanism

The text and image features are employed in conjunction to achieve optimal classification outcomes. This is because considering the semantic information of both image and text will enable the model to obtain more comprehensive contextual information. In addition, this will also enhance the generalization ability of the model so that the Crack-CLIP model can better understand and make correct judgments even when facing brand new samples. The output tensor ***C*** is mainly calculated by Equation (2) and (3).

$$C = T + I \qquad (2)$$
$$C_n = T_n + I_n \qquad (3)$$

## 3.2 Bayesian approach

In this study, we use a Bayesian neural network to enhance the uncertainty and adaptability of the model. The structure can be seen from Figure 4. The tensor ***C*** with shape [1,512] is passed through the Bayesian linear layer, the ReLU activation function, and the Dropout layer in order. Here the role of the Bayesian linear layer is to model the weights and biases so that they have the properties of normally distributed random variables. The Bayesian linear layer is invoked in the Pyro library.

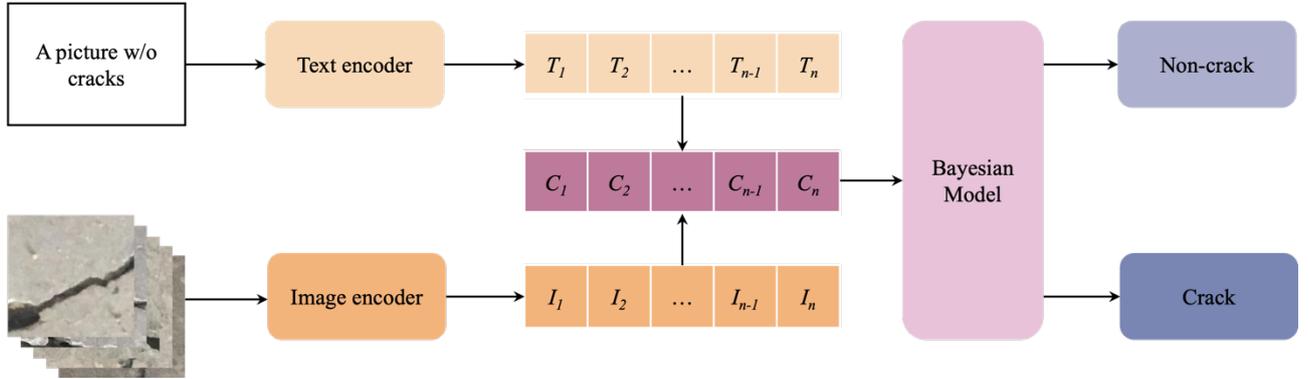

Figure 2: Architecture of Crack-CLIP

## 3.3 Evaluation metrics

The parameters in the CLIP section were all kept constant throughout the training of the model. In this study we used *Precision* (*P*), *Recall* (*R*), *F*1 Score and Precision-Recall Area Under the Curve (PR-AUC) to evaluate the Crack-CLIP model. The formulas are shown in Equations (4)-(6), respectively.

$$P = TP / (TP+FP) \qquad (4)$$
$$R = TP / (TP+FN) \qquad (5)$$
$$F1 = 2 \times P \times R / (P + R) \qquad (6)$$

where *TP* means the number of correctly predicted positive samples, *FP* is the number of incorrectly predicted negative samples, *FN* represents the number of incorrectly predicted positive samples. PR-AUC is the area under the curve of precision versus recall and is used to measure the performance of the model at different thresholds.

## 4 RESULTS AND DISSCUSSION

### 4.1 Results for datasets with different proportion

In this study, the normal linear layer was employed as our comparison to demonstrate the effectiveness of Bayesian networks in the CLIP model. The results of the few-shot learning classification for different proportions of datasets are shown in Table 2. All models were trained to convergence, and the losses during training are shown in Figure 4.

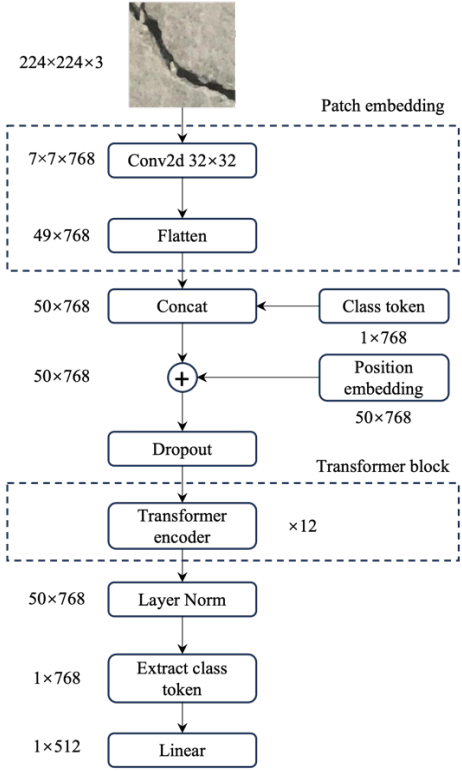

Figure 3: Image encoder

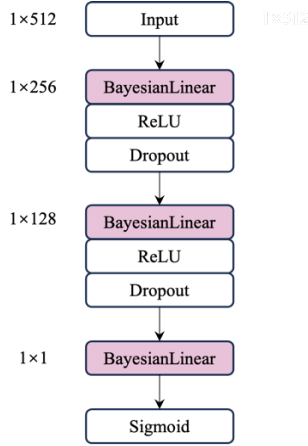

Figure 4: Bayesian model

Table 2 shows the effect of different proportions of datasets on the performance of the model when the CLIP model is subjected to few-shot learning. It can be seen from the results that as the proportion of the training set increases, the performance metrics of the model are significantly improved. The model's performance is inferior when it is employed directly for classification without any training. However, when trained with a limited number of samples (T1), the metrics demonstrate a notable enhancement. As the volume of data increases, the metrics demonstrate a gradual improvement. In particular, the model demonstrates high accuracy in recognizing positive class samples for both T4 and T5.

Table 2: Results for datasets with different proportion

| Number | $P$ | $R$ | $F1$ | PR-AUC |
|---|---|---|---|---|
| T0 | 0.9098 | 0.3761 | 0.5322 | 0.8477 |
| T1 | 0.9915 | 0.9882 | 0.9898 | 0.9995 |
| T2 | 0.9955 | 0.9963 | 0.9959 | 0.9999 |
| T3 | 0.9979 | 0.9980 | 0.9980 | 0.9999 |
| T4 | 0.9984 | 0.9983 | 0.9983 | 1.0000 |
| T5 | 0.9991 | 0.9986 | 0.9988 | 1.0000 |

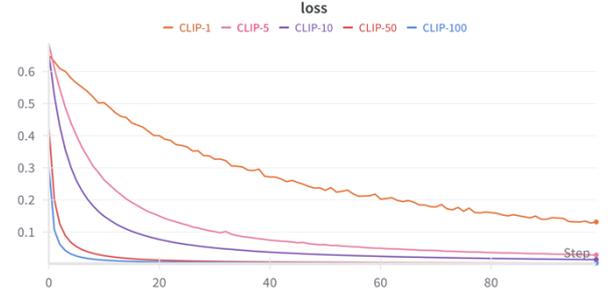

Figure 5: Bayesian model

These results indicate that few-shot learning is superior to zero-shot learning in practical engineering applications. In few-shot learning, the expansion of the dataset can markedly enhance the model's performance, thereby facilitating its capacity for generalization. Furthermore, the CLIP model exhibits a relatively high level of accuracy when only a limited number of samples are available, which is indicative of its proficiency in feature extraction and representation learning.

### 4.2 Results for Bayesian optimization

Table 3: Results for Bayesian optimization

| Number | $P$ | $R$ | $F1$ | PR-AUC |
|---|---|---|---|---|
| T1 | 0.9915 | 0.9882 | 0.9898 | 0.9995 |
| T1-B | 1.0000 | 1.0000 | 1.0000 | 1.0000 |
| T5 | 0.9991 | 0.9986 | 0.9988 | 1.0000 |
| T5-B | 1.0000 | 1.0000 | 1.0000 | 1.0000 |

Table 3 presents the performance results based on the Bayesian linear layer model. We can find that the Bayesian optimized model shows significant improvement in all the metrics compared to the unoptimized model. On the *Precision* metric, the optimized model achieves 100% accuracy on the test set, in comparison to 99.15% for the unoptimized model. The performance on the other three metrics is comparable. These results validate the enhancement of the Bayesian linear layer for the few-shot learning model.

The Crack-CLIP model demonstrated 100% classification accuracy in both the T1 training set, which comprised a smaller number of samples, and the T5 dataset, which included a greater number of samples. This is primarily attributable to the following factors: (1) The classification dataset used in this study is relatively simple, and the divergence between positive and negative samples is large, so the classification accuracy can reach 99% even without Bayesian linear layer. (2) The Bayesian linear layer can adapt to alterations in the data set by updating the probabilistic model in real time, thereby enabling the model to function effectively in a dynamic environment. This adaptability permits the model to be continuously optimized and to maintain high performance during the training process.

## 5 CONCLUSIONS

In this study, a crack image classification model combining CLIP and Bayesian optimization was successfully developed. The model displays excellent classification capabilities and has the potential for a wide range of applications in a few-shot learning scenario. This achievement not only improves the robustness and adaptability of the model, but also provides new solutions for future engineering applications. It is experimentally demonstrated that the model can maintain a high level of performance when dealing with practical cracks data. Future research can further explore the application of this method in other fields for wider generalization and application.

## ACKNOWLEDGEMENTS

The research is partly supported by the New Faculty Start-up Fund from City University of Hong Kong with grant number 9610612, partly supported by the Research Matching Grant Scheme with grant number 9229141, partly supported by the CityU Strategic Interdisciplinary Research Grant with grant number 7020076.